\documentclass[letterpaper, 10 pt, conference]{ieeeconf} 

\usepackage{graphicx}
\usepackage{amsmath}
\usepackage{algorithm}
\usepackage{algpseudocode}
\usepackage{booktabs}
\usepackage{multirow}
\usepackage{amssymb}
\usepackage{bbding}
\usepackage{pifont}
\usepackage{tabularx}
\usepackage{hyperref}
\usepackage{comment}
\hypersetup{
    colorlinks=true,
    linkcolor=magenta,
    filecolor=magenta,      
    urlcolor=magenta,
    citecolor=magenta,
}

\IEEEoverridecommandlockouts                              

\title{\LARGE \bf
LiloDriver: A Lifelong Learning Framework for Closed-loop Motion Planning in Long-tail Autonomous Driving Scenarios
}

\author{Huaiyuan Yao$^{*1}$, Pengfei Li$^{*2}$, Bu Jin$^{4}$, Yupeng Zheng$^{4}$, An Liu$^{3}$,\\ Lisen Mu$^{5}$, Qing Su$^{5}$, Qian Zhang$^{5}$, Yilun Chen$^{\dag 2}$, Peng Li$^{\dag 2}$  
\thanks{\textbf{* Equal Contribution}, \textbf{\dag \space Corresponding Author.}}
\thanks{$^{1}$ Xi'an Jiaotong University, China
        {\tt\small huaiyuanyao@gmail.com}}%
\thanks{$^{2}$ Institute for AI Industry Research (AIR), Tsinghua University, China,
        {\tt\small li-pf22@mails.tsinghua.edu.cn, \{chenyilun, lipeng\}@air.tsinghua.edu.cn}}%
\thanks{$^{3}$ Department of Computer Science and Technology, Tsinghua University, China,
{\tt\small la22@mails.tsinghua.edu.cn}}
\thanks{$^{4}$ Institute of Automation, Chinese Academy of Sciences, China, {\tt\small \{jinbu2022, zhengyupeng2022\}@ia.ac.cn}}%
\thanks{$^{5}$ Horizon Robotics, China, {\tt\small \{lisen.mu, jerry.su, qian01.zhang\}@horizon.cc}}%
}

\begin{document}

\maketitle
\thispagestyle{empty}
\pagestyle{empty}

\begin{abstract}
Recent advances in autonomous driving research have focused on developing motion planners that are robust, safe, and adaptive. However, existing rule-based and data-driven planners lack the adaptability required to handle long-tail scenarios, while knowledge-driven methods offer strong reasoning capabilities but face challenges in representation, control, and real-world evaluation. To address these challenges, we present LiloDriver, a lifelong learning framework for closed-loop motion planning in long-tail autonomous driving scenarios. By integrating large language models (LLMs) with a memory-augmented planner generation system, LiloDriver continuously adapts to new scenarios without retraining. It features a four-stage architecture including perception, scene encoding, memory-based strategy refinement, and LLM-guided reasoning. Evaluated on the nuPlan benchmark, LiloDriver achieves superior performance in both common and rare driving scenarios, outperforming static rule-based and learning-based planners. Our results highlight the effectiveness of combining structured memory and LLM reasoning to enable scalable, human-like motion planning in real-world autonomous driving. Our code is available at \href{https://anonymous.4open.science/r/LiloDriver-7ED6/}{https://anonymous.4open.science/r/LiloDriver-7ED6/}
\end{abstract}

\section{INTRODUCTION}
In autonomous driving systems, motion planning serves as a fundamental component that generates an optimal trajectory for the ego vehicle based on perception data, thereby ensuring safety, comfort, and driving efficiency~\cite{motionplanning, predictionplanning}. Recently, a notable closed-loop evaluation paradigm~\cite{nuplan} has advanced the field by emphasizing real-time performance in interactive simulations, rather than merely aligning with expert trajectories~\cite{rethinking_openloop_evaluation}.

\begin{figure}[!h]
    \centering
    \includegraphics[width=0.5\textwidth]{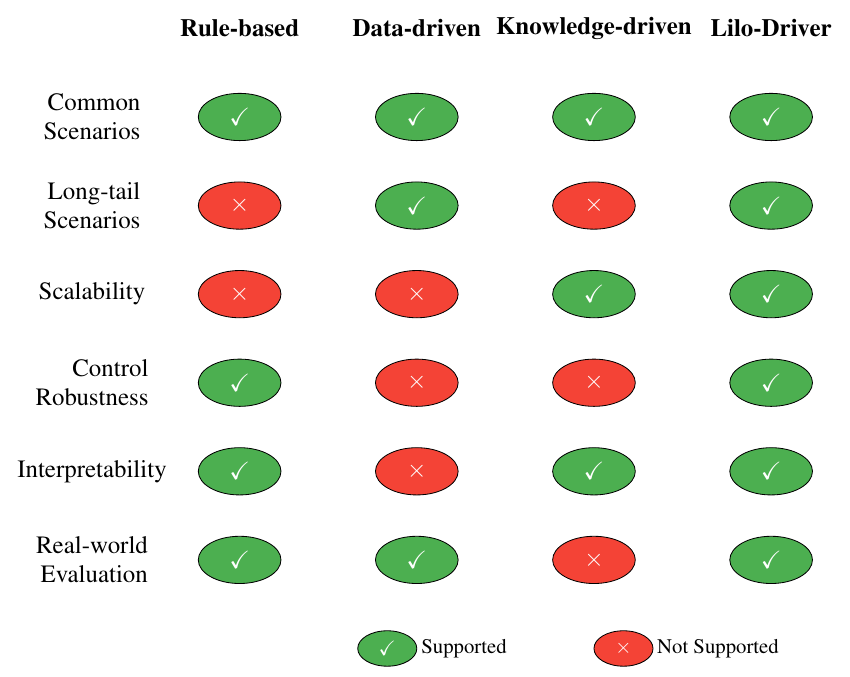}
    \caption{Comparison of four planning paradigms for autonomous driving across key criteria, including generalization to long-tail scenarios, scalability, robustness, interpretability, and real-world deployability. LiloDriver demonstrates comprehensive advantages over rule-based, data-driven, and knowledge-driven approaches.}
    \label{fig:comparsion}
\end{figure}

Existing end-to-end data-driven and rule-based systems have achieved remarkable results in common scenarios but struggle to generalize or adapt over time~\cite{llm4drivesurveylargelanguage, MLLMforAD, visionfoundationmodels, visionlanguagemodelsautonomous}, as illustrated in Fig.~\ref{fig:comparsion}. This is largely due to the inherently \textit{long-tail nature} of autonomous driving, where rare yet safety-critical situations are highly diverse and unpredictable~\cite{motionplanninggeneralize}. Static designs and fixed learning objectives limit current frameworks in handling such complex and evolving cases.
Rule-based methods generally utilize predefined heuristics to mimic human driving behavior. PDM-Closed~\cite{PDM}, which extends IDM~\cite{IDM} with pre-simulation, achieved state-of-the-art results in the NuPlan Challenge~\cite{nuplan}, showing strong performance in common scenarios. However, their rigidity limits generalization to rare and dynamic long-tail cases~\cite{lmmenhancedsafetycriticalscenariogeneration}. While generalization to unseen scenarios is essential for learning-based planners, recent studies show that they still suffer from overfitting, limited robustness, and poor transferability~\cite{planagent}. Once deployed, these models are static and require costly retraining to incorporate new knowledge, limiting their adaptability to dynamic environments and rare cases while risking the stability of previously validated behaviors.

In short, current closed-loop motion planning faces two core challenges: fixed strategies struggle to adapt to diverse long-tail scenarios, and retraining data-driven models for rare cases is costly and impractical. This motivates the need for a \textbf{lifelong learning} capability, enabling driving agents to continuously acquire new knowledge and adapt to unseen edge cases in closed-loop planning. Achieving such continual adaptation remains a key challenge in autonomous driving.

Recent advances in large language models (LLMs) exhibit a promising future for autonomous driving, particularly in scene understanding, decision-making, and interpretability~\cite{surveymultimodallargelanguage, endtoendautonomousdriving}.  Language models possess intrinsic common-sense knowledge and demonstrate a general understanding of the surrounding world, which enables them to identify complex real-world scenarios and adapt their driving behavior accordingly. Nonetheless, existing knowledge-driven approaches that rely on LLMs remain simplistic and face several critical challenges: (1) ineffective scene representation that hampers LLM performance (e.g., using too many coordinate tokens or omitting lane information); (2) directly generating trajectory waypoints from LLM outputs, which can lead to hallucinations and poor control robustness; and (3) a lack of systematic real-world evaluation, as these methods are typically tested only in simplified environments (e.g., Highway-env or SUMO).

To address these challenges, we propose LiloDriver, a lifelong learning framework for closed-loop motion planning in long-tail autonomous driving scenarios. Our system integrates LLM-based reasoning with dynamic knowledge augmentation via a memory bank, structured into a four-stage architecture: 
(1) \textbf{Environment and Perception} module constructs scene context from maps and agent histories, providing the basis for long-tail awareness; (2) \textbf{Scene Encoder} distills this context into latent representations that support generalization beyond seen examples; (3) \textbf{Memory and Planner Generation} module enables continuous adaptation by organizing past experiences and generating behavior strategies suited to novel scenarios; (4) \textbf{Reasoning and Execution} module leverages LLMs to compose planning decisions grounded in both current context and accumulated knowledge. This design enables LiLoDriver to dynamically assign planners, update strategies at inference time, and continually evolve through new experiences—mimicking human-like learning in the driving loop.

We evaluate LiloDriver on the real-world autonomous driving dataset nuPlan, with a focus on long-tail scenarios and closed-loop performance. Experimental results show that our method significantly outperforms both static planners and conventional learning-based approaches, especially in rare and challenging driving situations.

Our main contributions are summarized as follows:

\begin{itemize}
    \item To the best of our knowledge, we present the first lifelong learning paradigm for motion planning in long-tail autonomous driving scenarios, inspired by the way humans incrementally acquire and adapt driving skills.
    
    \item We propose LiloDriver, a novel LLM-based framework that supports this paradigm by combining memory-driven experience accumulation, inference-time decision-making via large language models, and scenario-specific planner generation.
    
    \item Extensive experiments demonstrate that LiloDriver enables continual adaptation and generalization to unseen long-tail scenarios without retraining, showcasing its effectiveness and potential for scalable deployment in real-world autonomous driving systems.
\end{itemize}

\begin{figure*}[htp]
  \centering
  \includegraphics[width=1.0\textwidth]{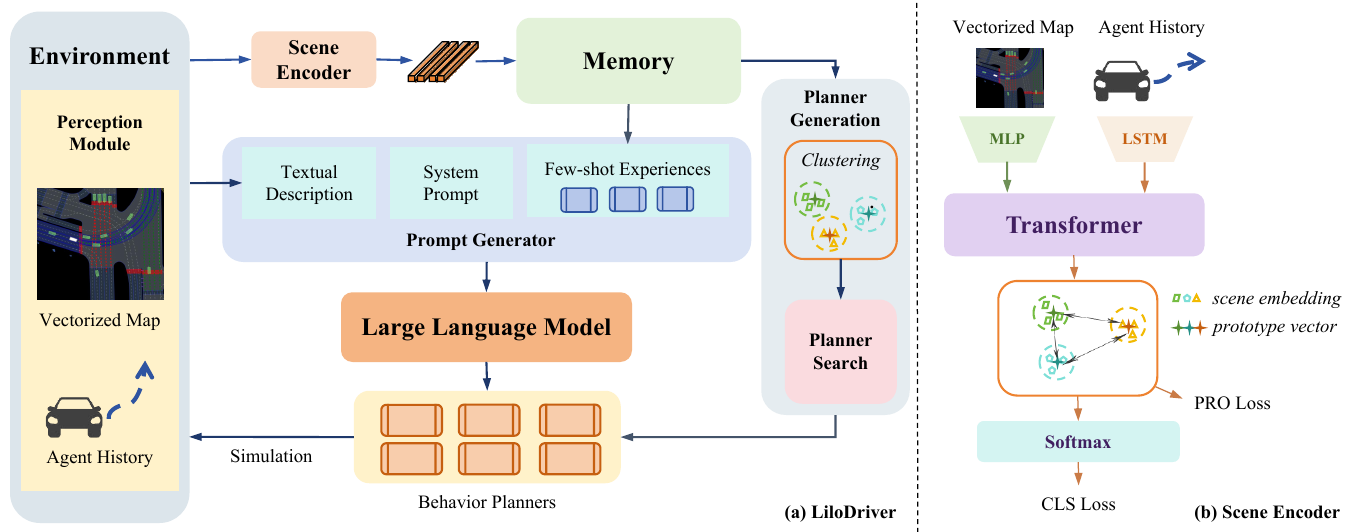}
  \caption{\textbf{The overall architecture of LiloDriver} comprises four core modules: (1) \textbf{Environment and Perception}, which integrates vectorized maps and agent histories to construct scene context; (2) \textbf{Scene Encoder}, which converts multi-modal perception inputs into latent embeddings for scene representation; (3) \textbf{Memory and Planner Generation}, which organizes clustered scene embeddings and associated few-shot planning experiences for planner adaptation; (4) \textbf{Reasoning and Execution}, which leverages an LLM to select appropriate behavior planners based on the current scenario.}
  \label{fig:pipeline}
\end{figure*}

\section{Related Work}

\subsection{Motion Planning in Autonomous Driving}
The field of autonomous driving has traditionally followed a modular framework consisting of perception, prediction, and planning components like ApolloICRA~\cite{baiduapollo}.  
In academia, the large-scale closed-loop benchmark nuPlan~\cite{nuplan} highly motivates autonomous driving research. PDM~\cite{PDM} extends IDM with different hyper-parameters to win the nuPlan challenges that showcases excellent control performance of rule-based models in trivial scenarios. However, PDM struggles to generalize to more complex, long-tail scenarios. 
Thus recent advancements in motion planning have focused on learning-based models to imitate expert behaviors. UrbanDriver~\cite{urbandriver} leverages PointNet-like architectures to reason globally about the driving environment. DTPP~\cite{dtpp} takes a differentiable approach, jointly training both the trajectory prediction model and cost function for enhanced performance in ego-conditioned scenarios. GameFormer~\cite{gameformer} uses a transformer-based encoder-decoder structure and frames planning as a level-k game. GC-PGP~\cite{gc-pgp} focuses on imitating expert drivers but lacks robustness in complex long-tail scenarios, while PlanTF~\cite{plantf} introduces a state dropout encoder. In PlanTF, a long-tail scenario benchmark \textit{Test14-hard} is proposed to validate model generalization in rare and complex situations. Our model builds on a modular paradigm within closed-loop settings. We combine the generalization capabilities of learning-based approaches with the robust control of rule-based methods to address long-tail scenarios.

\subsection{Large Language Models for Autonomous Driving}
The Large Language Model has proven highly effective in common-sense reasoning and zero-shot generalization. Building upon this foundation, the Multi-Modal LLM incorporates additional modality encoding, extending its capacity to process and understand multi-modal data~\cite{comal, da2025generativeaitransportationplanning}.

To enhance the driving knowledge of LLM, some research adopts prompt-tuning approaches such as SurrealDriver~\cite{surrealdriver}. DiLu~\cite{dilu} constructs a closed-loop LLM-based knowledge-driven driving model using the simple Highway-env environment. Furthermore, LanguageMPC~\cite{languagempc} is the first to develop a controller that translates LLM decisions into actionable driving commands. While LLM-Assist~\cite{llmassist} refines the hyperparameters of rule-based controllers, PlanAgent~\cite{planagent} directly generates planner parameters. Regarding fine-tuning LLMs for autonomous driving, LMDrive~\cite{lmdrive} has implemented the first LLM-based closed-loop end-to-end autonomous driving system based on the CARLA simulator. DriveMLM~\cite{drivemlm} introduces a novel framework that integrates LLMs with existing autonomous driving modules to take decision-making in real-world scenarios. In our approach, we encode the multi-modal perception inputs and employ a memory to improve its ability to understand driving scenarios.

\section{Methodology}

\subsection{Overview of the Architecture}

To address the challenges of closed-loop motion planning in long-tail scenarios, we propose LiloDriver, a lifelong learning framework that integrates LLM-based reasoning with dynamic knowledge adaptation. As illustrated in Fig.~\ref{fig:pipeline}, LiloDriver comprises four main components working in synergy. The Scene Encoder utilizes vectorized map information and historical agent trajectories from the Environment and Perception Module to create an environment-aware latent space for scene representation. The Memory stores diverse scene embeddings, while the Planner Generation Module clusters rare scenarios and optimizes planning strategies through grid search. As new situations are encountered, iterative memory updates enable continual adaptation and improved decision-making. Finally, the Reasoning and Execution Module leverages textual descriptions and few-shot examples retrieved from memory, enabling the LLM to generate contextually appropriate motion planning decisions. This architecture allows LiloDriver to dynamically refine its planning strategies at inference time, effectively mimicking human-like learning within the driving loop.

\subsection{Environment and Perception Module}
To enhance scenario comprehension efficiently, the Environment and Perception Module leverages static vectorized maps and dynamic agent histories as its perception inputs.

\subsubsection{Vectorized Map}
We utilize three key static elements to construct the map $M$: roads, crosswalks, and route lanes, within a query radius \( r \) centered on the ego vehicle. The encoder extracts map elements in polygonal format and converts them into feature tensors through linear interpolation. The road tensor \( M_R \in \mathbb{R}^{40 \times 50 \times 7} \) represents up to 40 lanes, each described by 50 centerline waypoints. Each waypoint \( M_w \in \mathbb{R}^{7} \) comprising the position coordinates, heading angle, and traffic light state. The crosswalk tensor  \( M_C \in \mathbb{R}^{5 \times 30 \times 3} \) and the route lane tensor \( M_L \in \mathbb{R}^{10 \times 50 \times 3} \) similarly encode their respective spatial features.

\subsubsection{Agent History}
Effective motion planning requires a comprehensive understanding of the dynamics of surrounding agents. The encoder tracks up to 
$N$ nearby entities including vehicles, pedestrians, and cyclists, to model their motion through historical trajectories over the past $T$ seconds. The resulting agent tensor \( H \in \mathbb{R}^{N \times T \times k} \) encodes $k$ attributes for each agent, such as position, velocity, yaw rate, and type. All features are normalized with respect to the ego vehicle’s current state.

\subsection{Scene Encoder}
To learn high-quality latent representations, we design a Scene Encoder and incorporate a hybrid loss function to improve the efficiency of scene representation learning. The encoder processes agent histories $H$ using an LSTM network, while vectorized map $M$ is encoded via a multi-layer perceptron (MLP) and subsequently aggregated through max-pooling. To capture spatial and contextual relationships among scene elements, the resulting agent-wise scene context tensor is further refined through a Transformer encoder. 

To define the loss function, we introduce the concept of \textit{prototype}, where each known class of driving situations is represented by a discriminative feature vector that serves as the core representation for that class. The encoder learns these prototypes using a novel loss function \( L_{PRO} \), which encourages the model to maximize inter-class separability while minimizing intra-class variance, thereby promoting robust feature discrimination.  

The loss function \( L_{PRO} \) is formulated as:  

\begin{align}
L_{PRO} &= y_{ij} \cdot \max(D_{ij} - m_p, 0)  \notag \\ 
        &\quad + \max \left( [1 - y_{ij}] \cdot \max(m_n - D_{ij}, 0) \right)
\end{align}
where
\begin{align}
D_{ij} = 1 - \frac{z_i \cdot P_j}{\|z_i\| \|P_j\|}
\end{align}

Here, \( y_{ij} \) is an indicator variable, where \( z_i \) corresponds to the embedding of the current scenario, and \( P_j \) represents the prototype of class \( P_j \). The loss function encourages proper clustering of embeddings by comparing the cosine distance between the scenario embedding \( z_i \) and the prototype \( P_j \).

To incorporate the classification results, we add the softmax classification loss \( L_{CLS} \). The final loss function is:

\begin{align}
    L = L_{PRO} + \lambda \cdot L_{CLS}
\end{align}
where \( \lambda \) is a hyperparameter that balances the two terms, ensuring that the goals of clustering and classification objectives are optimized during training.

\begin{algorithm}[htp]
\renewcommand{\arraystretch}{0.8}
    \caption{Clustering and Grid Search Optimization}
    \label{alg:dbscan_grid_search}
    \begin{algorithmic}[1]
        \Require Driving scenario embeddings \( S = \{s_1, s_2, ..., s_n\} \), clustering parameters \( \epsilon, minPts \), Planner parameters \( P \)
        \Ensure Optimized planning strategies for each cluster
        
        \State \textbf{Step 1: Scenario Clustering}
        \For{each scenario \( s_i \in S \)}
            \State Find neighbors \( N(s_i) = \{s_j \mid D(s_i, s_j) \leq \epsilon\} \)
            \If{\(|N(s_i)| \geq minPts\)}
                \State Assign \( s_i \) as a core point and expand cluster
            \ElsIf{\(s_i\) is reachable from a core point}
                \State Assign \( s_i \) as a border point
            \Else
                \State Mark \( s_i \) as noise
            \EndIf
        \EndFor
        
        \State \textbf{Step 2: Planner Search Optimization}
        \For{each cluster \( C_k \)}
            \For{each parameter set \( P_m \) in grid search space}
                \State Apply planner \( P_m \) to scenarios in \( C_k \)
                \State Evaluate performance by simulation \( L(C_k, P_m) \)
            \EndFor
            \State Select optimal planner parameters:
            \[
            P_k^* = \arg\max_{P_m} L(C_k, P_m)
            \]
        \EndFor
    \end{algorithmic}
\end{algorithm}

\subsection{Memory and Planner Generation}

To ensure continuous adaptability and refinement of motion planning strategies, LiloDriver incorporates a memory-based approach for lifelong learning. This mechanism allows the system to evolve over time by storing, clustering, and optimizing planning strategies across diverse driving scenarios.

In the memory module, each driving scenario is stored as an embedding vector in the latent space. Memory serves as a repository for all previously encountered driving situations, allowing LiloDriver to access these past experiences when making decisions for new situations. The embedding of each scenario is indexed and stored with associated metadata, such as scenario labels and planning strategies.

In the subsequent Planner Generation, LiloDriver design a clustering method based on DBSCAN~\cite{dbscan}.This unsupervised approach groups similar scenarios together based on their proximity in the latent space. The algorithm clusters driving scenarios by identifying core points with a minimum number of neighbors within a specified radius. It then expands clusters by including neighboring points that are density-connected. Points that don’t meet the density requirement are labeled as noise and are not assigned to any cluster. This clustering process is incrementally updated as new scenarios are encountered, allowing LiloDriver to grow and adapt its memory over time.

Once the scenarios are clustered, LiloDriver conducts a planner search to identify the optimal behavior planner parameters, following the approach in~\cite{PDM}. For each cluster, the search explores a grid of candidate parameters—such as the minimum gap to the leading agent and the maximum acceleration—to select the most suitable configuration. This process ensures that the behavior planners are fine-tuned for specific scenario clusters, enhancing the robustness and adaptability of motion planning under long-tail conditions.

\begin{figure*}[t!]
  \centering
  \includegraphics[width=0.8\textwidth]{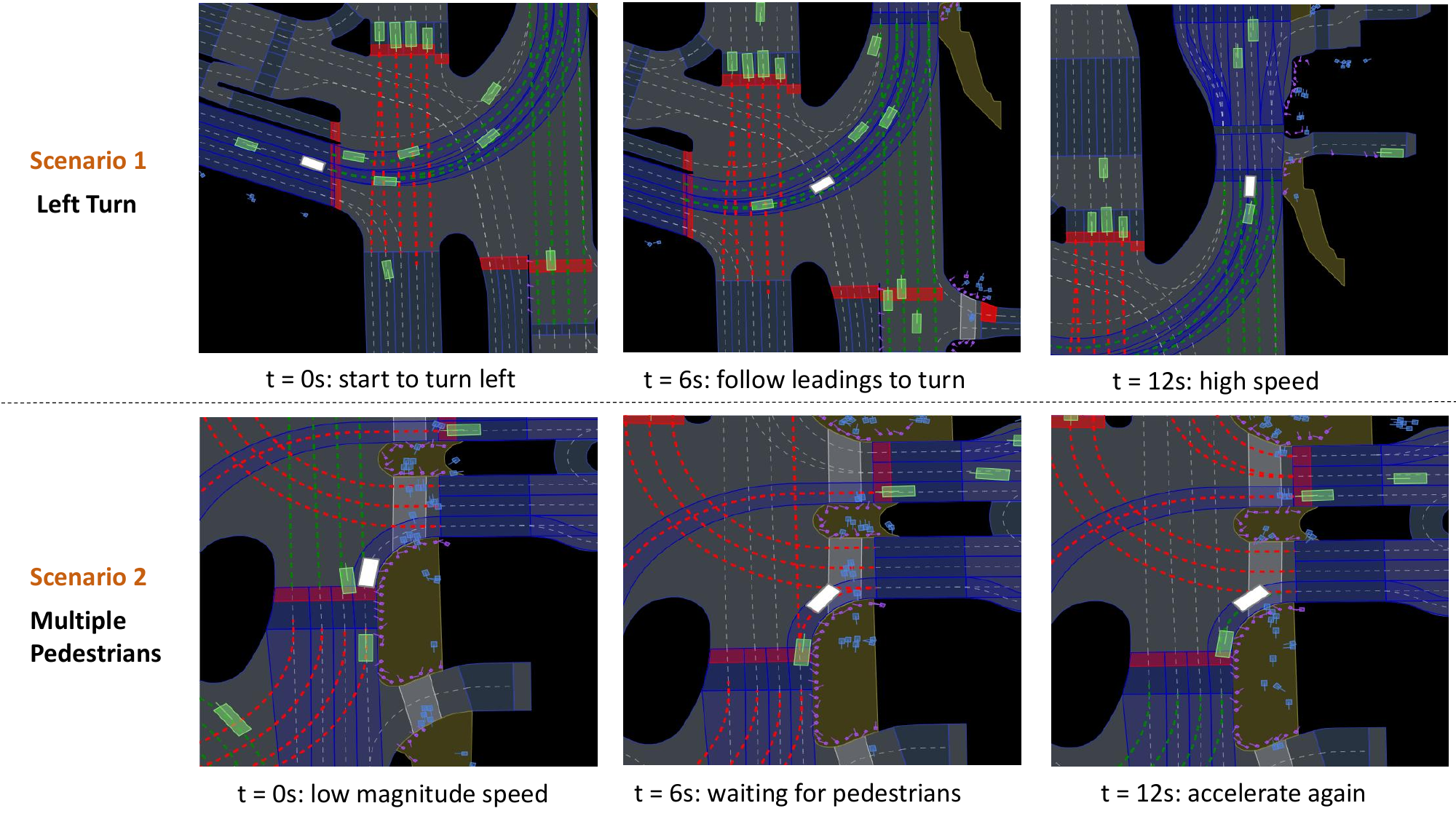}
  \caption{The demonstration of LiloDriver in real-world long-tail scenarios. The first row illustrates a left-turning behavior where the vehicle smoothly adjusts its trajectory over time. The second row shows a pedestrian-rich environment where LiloDriver exhibits cautious and adaptive planning by slowing down, yielding, and resuming motion. This highlights the system’s ability to handle complex, dynamic traffic conditions with human-like decision-making.}
  \label{fig:vis}
\end{figure*}

\subsection{Reasoning and Execution Module}
The Reasoning module receives scene features and prompts from the Perception module. LiloDriver first uses a prompt generator to convert scene features into textual prompts, which is then fed into the LLM. Guided by a chain-of-thought approach, the LLM interprets the driving scenario using these embeddings along with few-shot examples from the memory, determines the appropriate driving behavior.

\textbf{Text Description}  
To improve scenario comprehension and leverage the inherent common-sense reasoning of LLMs, we enrich the context with a system prompt, motion description, and a chain of thought. The system prompt provides a clear summary of the closed-loop driving task and outlines the inputs and outputs for the task. The LLM is guided through a chain of thought, breaking down the task and reasoning step-by-step. The motion description captures the dynamics of the ego vehicle and other agents in the environment. LiloDriver leverages this descriptor to construct a natural language prompt based on the scene features, which the LLM then uses to infer and generate appropriate driving behaviors for different traffic conditions.

\textbf{Behavior Planners}  
Real-world drivers display a wide range of behaviors, from conservative and cautious driving to assertive and reckless maneuvers. Planning methods must be capable of handling these diverse conditions. Through the memory module, the LLM agent learns to adopt an appropriate driving style based on similar scenarios and the corresponding grid search results. Each driving behavior is executed by a rule-based behavior planner.

\begin{equation}
    \mathrm{DrivingBehaviorPlanner}(lo, s_0, a_m, b, th)
\end{equation}

Behavior planners run at high frequency, while LLM reasoning is queried over a longer time frame. We fine-tune the PDM-Closed model using conditioned parameters to function as behavior planners that execute corresponding driving actions. Similarly, our PDM-based behavior planner extends the IDM by incorporating forecasts and pre-simulations to select the best IDM planner within lateral offsets \( lo \) and speed limit \( v_0 \). Given parameters such as maximum acceleration \( a_m \), maximum deceleration \( b \), and minimum gap to the leading agent \( s_0 \), the IDM planner calculates longitudinal acceleration using the following equation:

\begin{equation}
    a = a_{m} \left[1-\left(\frac{v}{v_0}\right)^\delta - \left(\frac{s^*}{s}\right)^2\right]
\end{equation}
\begin{equation}
    s^* (v,\Delta v) = s_0 + \max \left(0, vT+\frac{v\Delta v}{2 \sqrt{a_m b}}\right)
\end{equation}

The target speed \( v_0 \) is set according to the speed limits of individual roads, while braking decisions rely on a threshold \( th \) calculated from the real-time time-to-collision (TTC) value. 

\begin{table}[htp]
\renewcommand{\arraystretch}{1.0}
\caption{Closed-loop Metrics of AdaptDriver on NuPlan Benchmark}
\label{table:main}
\begin{center}
\label{table:main}
\resizebox{\columnwidth}{!}{
\begin{tabular}{llcc}
\toprule
Category & Planning Models & Val14-split & Test14-hard \\
\midrule
Expert & Log-replay & 94.03 & 85.96 \\
\midrule
\multirow{2}{*}{Rule-based} & IDM~\cite{IDM}  & 70.39 & 56.16 \\
 & PDM-Closed~\cite{PDM}  & 92.81 & 65.07 \\
\midrule
\multirow{7}{*}{Data-driven} & RasterModel~\cite{nuplan}  & 69.66 & 49.47 \\
 & UrbanDriver~\cite{urbandriver} & 63.27 & 51.54 \\
 & GC-PGP~\cite{gc-pgp}  & 55.99 & 43.22 \\
 & PDM-Open~\cite{PDM} & 52.80 & 33.51 \\
 & GameFormer~\cite{gameformer}  & 80.80 & 66.59 \\
 & PlanTF~\cite{plantf} & 84.83 & 72.68 \\
 & DTPP~\cite{dtpp}  & 89.64 & 59.44 \\
\midrule
\multirow{3}{*}{Knowledge-driven} & LLM-ASSIST (UNC)~\cite{llmassist}  & 90.11 & - \\
 & LLM-ASSIST (PAR)~\cite{llmassist}  & 93.05 & - \\
 & PlanAgent~\cite{planagent}  & 93.26 & 72.51 \\
\midrule
\multirow{2}{*}{Lifelong Learning} & LiloDriver (Initial)  & 93.09 & 69.92 \\
 & LiloDriver (Evolved) & \textbf{93.33} & \textbf{73.10} \\
\bottomrule
\end{tabular}
}
\end{center}
\end{table}

\begin{table*}[htp]
\centering
\caption{
Lifelong learning performance on \textbf{Test14-Hard} via incremental memory augmentation. 
Following the \textbf{Offline Pre-training} of the scene encoder, the model sequentially encounters and adapts to four rare scenarios during test-time. 
Results demonstrate that LiloDriver effectively acquires new skills (noted by $\uparrow$) while maintaining stable performance on \textbf{Common} scenarios, showing no catastrophic forgetting.
\footnotesize
Abbreviations: \textbf{W}: Waiting for pedestrian; \textbf{N}: Near multiple vehicles; \textbf{C}: Changing lane; \textbf{T}: Traversing pickup/dropoff.
}
\label{table:ttt_lifelong}
\renewcommand{\arraystretch}{1.1} 
\begin{tabular}{ll c cccccc}
\toprule
\multirow{2}{*}{\textbf{Phase}} & \multirow{2}{*}{\textbf{Learning State}} & \textbf{Incremental} & \multicolumn{6}{c}{\textbf{Evaluation Score}} \\
\cmidrule(lr){4-9}
& & \textbf{Knowledge} & \textbf{Common} & \textbf{W} & \textbf{N} & \textbf{C} & \textbf{T} & \textbf{Total} \\
\midrule
\textit{Offline} & 1. Pre-trained Encoder & \textit{None} & 75.42 & 42.38 & 67.59 & 56.99 & 44.40 & 69.92 \\
\midrule
\multirow{4}{*}{\shortstack{\textit{Online}\\\textit{Lifelong}}} 
& 2. Encounter Scenario W & + \{W\} & 75.58 & \textbf{45.04}$\uparrow$ & 67.63 & 57.13 & 44.42 & 70.61 \\
& 3. Encounter Scenario N & + \{N\} & 75.45 & 46.23 & \textbf{68.59}$\uparrow$ & 56.42 & 44.84 & 71.37 \\
& 4. Encounter Scenario C & + \{C\} & 75.62 & 46.65 & 68.29 & \textbf{63.27}$\uparrow$ & 44.50 & 72.87 \\
& 5. Encounter Scenario T & + \{T\} & \textbf{75.65} & 46.04 & 68.63 & 64.54 & \textbf{47.45}$\uparrow$ & \textbf{73.10} \\
\bottomrule
\end{tabular}
\end{table*}

\section{Experiments}

In this section, we evaluate LiloDriver on the large-scale closed-loop nuPlan benchmark to validate its effectiveness in long-tail autonomous driving scenarios. We first describe the implementation details and evaluation protocol, followed by benchmark comparisons, continual learning analysis, and ablation studies. All reported results are based on closed-loop simulation without retraining the Scene Encoder during lifelong adaptation.

\subsection{Implementation Details}
\textbf{LLM Backbone.} The reasoning module is instantiated using the LLaMA-7B model~\cite{llama}, which serves as a high-level decision-making agent. The LLM operates at a lower frequency ($T=15$\,s) compared to the behavior planners (10\,Hz), enabling a decoupled architecture where high-level reasoning and low-level control are separated. This design ensures that language-based reasoning does not interfere with real-time trajectory execution.

\textbf{Scenario Datasets.} We conduct experiments on the nuPlan benchmark~\cite{nuplan}, a large-scale closed-loop dataset containing over 1,500 hours of real-world driving data collected from four cities: Boston, Pittsburgh, Las Vegas, and Singapore. nuPlan is designed to evaluate autonomous driving systems under diverse and challenging traffic conditions. To evaluate our model, we use the official nuPlan Val14-split for general performance assessment and the nuPlan Test14-hard set to focus on long-tail scenarios. Additionally, we sample 12,000 scenarios from the nuPlan training split to construct the training set for the Scene Encoder.

\textbf{Closed-loop Evaluation} We adopt the nonreactive closed-loop score provided by nuPlan as the evaluation metric for driving performance in simulation. It assesses traffic rule compliance, similarity to human driving behavior, vehicle dynamics, and goal achievement. Scores range from 0 to 100, with higher values indicating superior performance.

\subsection{Benchmark Performance}

Table~\ref{table:main} presents the closed-loop performance comparison on Nuplan benchmark. We compare LiloDriver against rule-based, data-driven, and knowledge-driven planners.

Rule-based methods such as IDM and PDM-Closed achieve strong performance in common scenarios but exhibit noticeable degradation under long-tail conditions. Learning-based planners improve generalization but still suffer from performance drops in rare and complex scenarios. Knowledge-driven methods narrow this gap but lack structured lifelong adaptation. LiloDriver achieves competitive performance on Val14-split while maintaining strong robustness on Test14-hard. Importantly, it is the only paradigm that preserves high performance across both common and long-tail settings without retraining model parameters. This demonstrates the advantage of combining structured memory augmentation with LLM-based reasoning for inference-time adaptation. The balanced performance across both benchmarks indicates that LiloDriver does not overfit to either frequent or rare scenarios, but instead maintains stable closed-loop behavior through dynamic planner assignment.


\subsection{Lifelong Learning Ability}
To evaluate the continual adaptation of LiloDriver under long-tail conditions, we conduct a continual learning experiment on the \textbf{Test14-Hard} benchmark. We incrementally introduce previously unseen rare scenarios into the memory while keeping the Scene Encoder fixed, simulating a real-world deployment where the vehicle encounters novel challenges over time. The results are summarized in Table~\ref{table:ttt_lifelong}.

We begin with an \textit{Offline} pre-trained model as a baseline, which was exposed only to common driving scenarios. This initial state achieves a total score of 69.92. As the model transitions to the \textit{Online Lifelong} phase, it sequentially encounters and incorporates specialized knowledge for four rare scenario types: waiting for pedestrians (W), maneuvering near multiple vehicles (N), changing lanes (C), and traversing pickup/dropoff areas (T). These four categories are specifically selected because they involve intricate dynamic interactions with other road agents and initially yield low performance scores, representing significant challenges for conventional planners.

The results demonstrate two key strengths of the LiloDriver framework. First, the framework achieves continuous performance gain. With the incremental addition of scenario-specific knowledge, the performance in each corresponding category improves significantly (denoted by $\uparrow$). For instance, the score for lane changing (C) jumps from 56.99 to 63.27 upon the inclusion of relevant memory, contributing to a final total score of 73.10. Furthermore, the system exhibits strong resistance to catastrophic forgetting. Notably, as the model acquires new skills for long-tail scenarios, its performance on \textit{Common} scenarios remains remarkably stable. This indicates that the memory-augmented approach allows for the expansion of the system's capability boundaries without degrading its existing fundamental driving skills.

\subsection{Safety Analysis}

\begin{table}[htbp]
\centering
\caption{Safety Performance Improvement on \textbf{Test14-Hard}. The results demonstrate that the overall score enhancement is primarily driven by significant improvements in safety-critical metrics ($1 - \text{Collisions}$) and Time-to-Collision (TTC) scores.}
\begin{tabular}{lccc}
\toprule
\textbf{Model Version} & \textbf{Collisions Rate $\downarrow$} & \textbf{TTC $\uparrow$} & \textbf{Total Score $\uparrow$} \\
\midrule
LiloDriver (Initial) & 12.45\% & 62.18 & 52.84 \\
LiloDriver (Evolved) & \textbf{4.12\%} & \textbf{78.45} & \textbf{56.69} \\
\bottomrule
\end{tabular}
\label{table:safety_metrics}
\end{table}

As illustrated in Table~\ref{table:safety_metrics}, the substantial growth in the total score is predominantly attributed to the mitigation of safety risks. Specifically, the collision rate drops from 12.45\% to 4.12\%, while the Time-to-Collision (TTC) metric improves significantly from 62.18 to 78.45. This observation aligns with the intrinsic nature of long-tail scenarios: unlike common driving tasks where efficiency is the primary differentiator, rare and edge-case scenarios are characterized by high-risk dynamics where safety—rather than mere progress—is the bottleneck for performance. By effectively retrieving historical lessons from the memory bank, LiloDriver (Evolved) demonstrates a superior ability to anticipate potential hazards and maintain a safer buffer in these critical situations, thereby ensuring robust closed-loop planning.

\subsection{Efficiency}
\subsubsection{Algorithm and Complexity}
Algorithm~\ref{alg:dbscan_grid_search} facilitates strategy evolution by first categorizing long-tail experiences into "scenario archetypes" via clustering, then identifying optimal planner parameters $P_k^*$ through automated grid search and closed-loop simulation for each cluster. This hierarchical approach optimizes computational efficiency; the clustering phase requires $O(N \log N)$ time with spatial indexing, while the optimization phase complexity $O(K \cdot M \cdot T_{sim})$ is minimized as the number of clusters $K$ is much smaller than the total scenarios $N$. The space complexity remains a manageable $O(N \cdot D)$ for scenario buffer maintenance.

\subsubsection{System Efficiency}
As detailed in Table~\ref{table:efficiency}, LiloDriver achieves real-time viability by decoupling high-frequency tactical execution from long-horizon strategic reasoning. This architecture manages latency effectively: the LLM provides cognitive guidance with a 0.4 s inference delay, while memory retrieval and updates are completed within 0.3 s. By bridging the gap between slow-thinking reasoning and fast-acting physical execution, the framework ensures robust performance within the computational and safety-critical constraints of real-world deployment.

\begin{table}[htbp]
\centering
\caption{System Efficiency and Temporal Configuration. The framework decouples high-frequency tactical execution from long-horizon strategic reasoning.}
\resizebox{\columnwidth}{!}{
\begin{tabular}{llc}
\toprule
\textbf{Category} & \textbf{Component} & \textbf{Value} \\
\midrule
\multirow{2}{*}{\textbf{Control Frequency}} & Behavior Planner Execution & 10 Hz \\
& Reasoning Query Horizon & 15.0 s \\
\midrule
\multirow{3}{*}{\textbf{Latent Latency}} & LLM Inference & 0.4 s \\
& Memory Retrieval & 0.1 s \\
& Memory Update & 0.2 s \\
\bottomrule
\end{tabular}
}
\label{table:efficiency}
\end{table}

\subsection{Ablation Study}
\begin{table}[htp]
\centering
\caption{Ablation study on LiloDriver.}
\label{table:ablation}
\begin{tabular}{ccccc}
\toprule
\textbf{No.} & \textbf{LLM} & \textbf{Memory} & \textbf{Scene Encoder} & \textbf{Test14-Hard} \\
\midrule
1 & \ding{55} & \ding{51} & \ding{51} & 69.19 \\
2 & \ding{51} & \ding{55} & \ding{51} & 69.04 \\
3 & \ding{51} & \ding{51} & \ding{55} & 71.20 \\
4 & \ding{51} & \ding{51} & \ding{51} & 73.10 \\
\bottomrule
\end{tabular}
\end{table}


We further analyze the contribution of each component in LiloDriver: LLM reasoning, memory augmentation, and Scene Encoder. The results are summarized in Table~\ref{table:ablation}.

Removing any single component leads to performance degradation. Without the LLM, the system lacks high-level contextual reasoning. Without memory, the planner cannot perform incremental adaptation to unseen scenarios. Without the Scene Encoder, structured scene representation is lost, reducing clustering effectiveness. The full system achieves the best performance, confirming that closed-loop adaptation emerges from the interaction between structured representation learning, memory-based retrieval, and language-guided planner selection. 

Together, these results validate the design of LiloDriver as a unified lifelong learning framework for robust closed-loop motion planning.

\section{Conclusion}

In this work, we present LiloDriver, a lifelong learning framework for closed-loop motion planning in long-tail autonomous driving scenarios. By integrating structured scene embeddings, density-based memory clustering, and LLM-guided planner selection, LiloDriver enables inference-time adaptation without retraining model parameters. Experimental results on the nuPlan benchmark demonstrate that our method maintains competitive performance in common scenarios while progressively improving robustness in rare and challenging cases. The incremental memory augmentation mechanism allows scenario-specific planner refinement, achieving continual performance gains with minimal additional supervision. Ablation studies further confirm that the synergy between representation learning, memory-based retrieval, and language-guided reasoning is essential for stable and adaptive closed-loop behavior.

Despite these promising results, several challenges remain. First, the current memory module operates primarily on structured embeddings and predefined planner parameter spaces; extending it toward richer multi-modal memory representations may further enhance scenario understanding and cross-domain generalization. Second, while LLM-based reasoning provides high-level adaptability, the interaction between language-guided decisions and low-level control policies can be further optimized to improve temporal consistency and safety guarantees. Future work will explore tighter coupling between symbolic reasoning and continuous control, more scalable memory organization mechanisms, and improved alignment between high-level intent generation and trajectory optimization. We believe these directions will contribute toward more robust, interpretable, and scalable lifelong autonomous driving systems.





\end{document}